\PassOptionsToPackage{dvipsnames}{xcolor}
\PassOptionsToPackage{numbers,compress}{natbib}
\documentclass[11pt,letterpaper]{our}

\usepackage{subcaption}
\usepackage{tcolorbox}
\tcbuselibrary{skins}
\usepackage{multirow}
\usepackage{array}
\usepackage{nicefrac}
\usepackage{bm}
\usepackage{mathtools}
\usepackage{cancel}
\usepackage{wrapfig}
\usepackage{afterpage}
\usepackage{algorithm}
\usepackage{algpseudocode}
\usepackage{mdframed}

\usepackage[numbers,compress]{natbib}
\newcommand{\bench}{\textbf{ActiveVision}}
\newcommand{\task}[1]{\emph{#1}}

\title{An Exam for Active Observers}
\author{Jiarui Zhang\textsuperscript{*}, Muzi Tao\textsuperscript{*}, Shangshang Wang\textsuperscript{*}, Ollie Liu, Xuezhe Ma, Willie Neiswanger\\
University of Southern California\\
\textsuperscript{*}Equal contribution}
\correspondingauthor{Jiarui Zhang, \href{mailto:jzhang37@usc.edu}{jzhang37@usc.edu}; Willie Neiswanger, \href{mailto:neiswang@usc.edu}{neiswang@usc.edu}}

\begin{document}

\begin{abstract}
Human vision is a closed loop: gaze is continuously redirected by intermediate hypotheses rather than a single snapshot. Decades of psychophysics and cognitive science have argued that this \emph{active observation} is essential for a wide range of tasks. Whether today's multimodal large language models (MLLMs) exercise active observation is an empirical question that current vision-language benchmarks do not answer. We introduce \bench{}, a benchmark that makes active observation measurable for MLLMs, comprising 17 tasks across 3 categories. Tasks are designed to force repeated visual perception rather than a single static description. Frontier MLLMs collapse on \bench{}: the highest-scoring model we evaluate, GPT-5.5 at the highest exposed reasoning-effort tier, solves only 10.6\% of items and scores zero on 11 of the 17 tasks, and even Claude Fable~5, despite topping most reasoning and coding leaderboards, solves just 3.5\%, far behind three human participants who average 96.1\%. Furthermore, much of the gap persists even when models write and run their own vision code. Such code is unreliable on realistic imagery, and catching its failures itself requires the active perception the models lack. Together, these results indicate that current MLLMs lack robust active visual observation, motivating architectures and training objectives that close the perception--reasoning loop.

\par\vspace{8pt}
{\centering
\small\normalfont\bfseries
\href{https://activevision.dev}{Website}
\enspace|\enspace
\href{https://github.com/saccharomycetes/ActiveVision}{GitHub}
\enspace|\enspace
\href{https://huggingface.co/datasets/activevisionai/ActiveVision}{Dataset}
\par}
\end{abstract}

\maketitle

\begin{figure}[H]
  \setlength{\abovecaptionskip}{3pt}
  \setlength{\belowcaptionskip}{-4pt}
  \centering
  \vspace{-2mm}
  \includegraphics[width=0.8\linewidth]{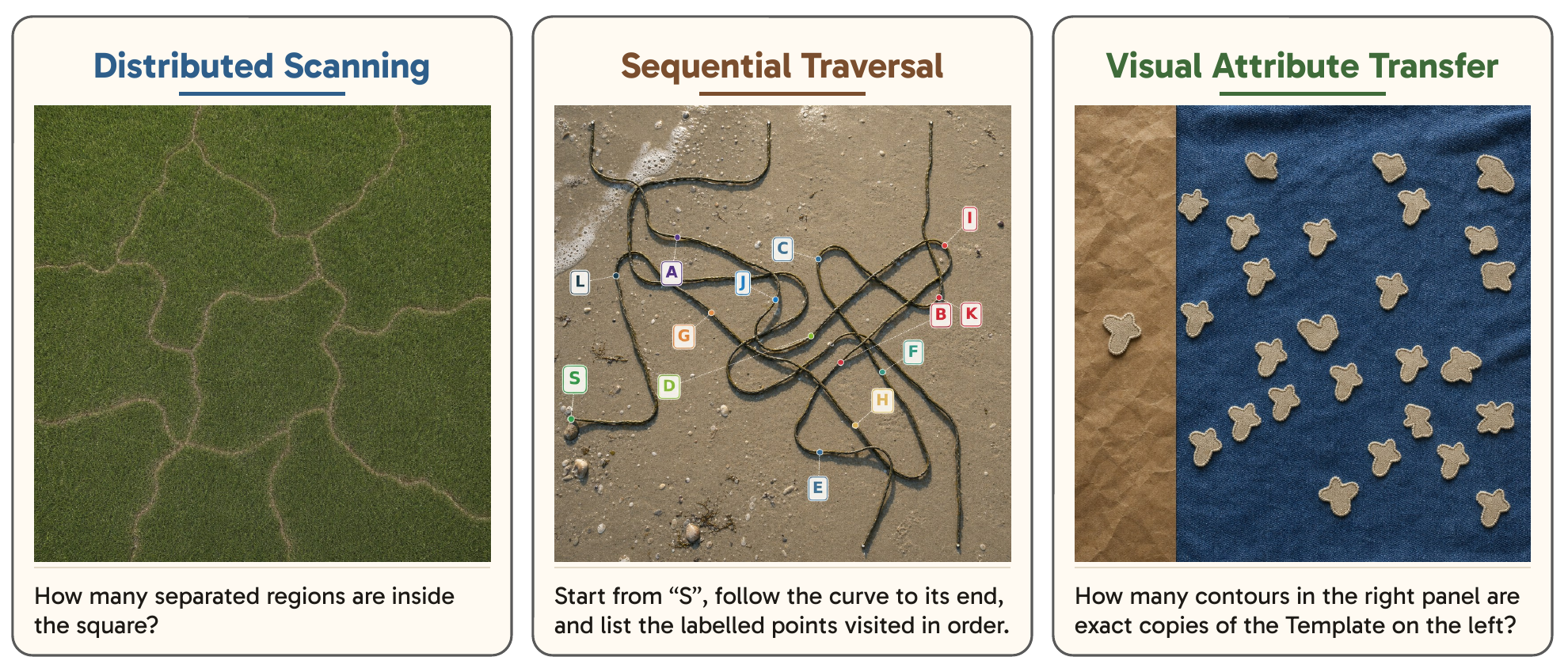}
  \caption{\textbf{Seeing is not always a single-glance task.} Many visual problems require an observer to coordinate attention and memory across an image: scanning exhaustively for distributed evidence, following connected structures without losing track, and comparing fine-grained attributes across distant regions. \bench{} turns these abilities into a controlled test of whether MLLMs can keep visual evidence in the reasoning loop.}
  \label{fig:teaser}
\end{figure}

\section{Introduction}
\label{sec:intro}

Try the three tasks in Figure~\ref{fig:teaser}. Count the separated regions of grassland in the aerial photo on the left. In the middle, trace the rope from the green ``S'' through all crossings to the far end, reading off each labeled point that you pass.
On the right, count every beige fabric patch on the linen whose silhouette matches the one shown on the kraft-paper template.
None can be solved at a glance.
Each requires several seconds of iterative, hypothesis-driven inspection: fixate, predict, return. The visual system executes this loop effortlessly, and the conscious mind barely notices it. Decades of research in psychophysics, cognitive science, and computer vision have argued that this iterative looking, known as \emph{active observation}, is not an ornament on top of perception but, for a wide range of tasks, an essential part of perception itself~\citep{active_vision,active_perception,animate_vision,findlay_gilchrist,hayhoe_natural,sensorimotor,yarbus}.

Multimodal large language models (MLLMs) have advanced rapidly, yet it remains unclear whether their benchmark gains reflect progress in this capability. From the GPT-4 family~\citep{gpt4} onward, frontier models such as Claude Fable~5~\citep{fable5}, GPT-5.5~\citep{gpt55}, and Gemini~3.1 Pro~\citep{gemini31} have achieved strong results on benchmarks commonly used to distinguish among frontier models. Performance on multimodal benchmarks such as MMMU-Pro~\citep{mmmu_pro} and CharXiv~\citep{charxiv} is nearing saturation, leaving little separation among frontier models.

Yet benchmark saturation does not mean that vision-language understanding is solved. Applications in robotics~\citep{robobench}, design~\citep{cadrille}, manufacturing~\citep{forge,benchcad}, computer use~\citep{osworld2}, spatial understanding~\citep{blueprintbench}, and scientific discovery~\citep{mac_bench,labbench} require models to revisit visual evidence as they reason, a capability current MLLMs do not reliably exhibit. Existing evaluations leave this requirement largely unmeasured. Public leaderboards remain dominated by captioning, single-image visual question answering, and multiple-choice tasks that can often be answered from a static description of the image. Few benchmarks assess whether models can revisit visual evidence to form, test, and refine hypotheses as reasoning unfolds.

Current MLLMs appear to be passive perceivers: the image is encoded once as a fixed sequence of visual tokens, with no explicit perception--action loop. Yet this need not preclude active observation. During autoregressive reasoning, the model may shift attention among visual tokens as hypotheses evolve, letting earlier visual findings guide later steps~\citep{zhang2025mllms}. The question is therefore behavioral rather than architectural:

\vspace{-1.5mm}
\begin{center}
\begin{tcolorbox}[
  enhanced,
  frame hidden,
  boxrule=0pt,
  arc=1pt,
  colback=CalGoldHex,
  borderline west={2.5pt}{0pt}{BerkeleyBlue},
  boxsep=0pt,
  left=6pt, right=6pt, top=7pt, bottom=7pt,
  before skip=2pt, after skip=2pt,
  halign=center
]
\small\textit{Do MLLMs perform \emph{active observation}, returning to the image to form and test hypotheses as they reason?}
\end{tcolorbox}
\end{center}
\vspace{-2mm}

We answer the question with \bench{}, a benchmark that makes iterative visual perception measurable. It consists of 17 tasks spanning three cognitive demands of active vision: \textbf{Distributed Scanning} for exhaustive coverage of many local signals, \textbf{Sequential Traversal} for ordered stepping along a connected structure, and \textbf{Visual Attribute Transfer} for fine-grained comparison across regions. Across these tasks, the underlying configurations resist concise language description, so a model cannot easily summarize an image once in language and answer from that summary alone; it must keep returning to the pixels as its reasoning unfolds.

To create this structural complexity and render it realistically, each task follows a two-stage construction pipeline. First, a procedural scaffold specifies the underlying geometry exactly, including arbitrary positions, shapes, and curves. Second, GPT-image-2~\citep{gptimage2} re-renders the scaffold as a photorealistic image while preserving positions, counts, labels, and topology (\S\ref{sec:tasks}). Voronoi regions become aerial fields and rivers, arrow chains become stones linked by footprints, and tangled loops become ropes on driftwood. The resulting images introduce realistic textures and visual complexity absent from sparse, cartoon-like renderings.

Every frontier model we evaluate fails on the overwhelming majority of the 85 released items. We test each model at every reasoning-effort tier exposed by its API. Across all settings, the best result is 10.6\%, achieved by GPT-5.5 \emph{xhigh}, which still scores zero on 11 of the 17 tasks. Claude Fable~5, despite topping most reasoning and coding leaderboards, solves just 3.5\%. Three human participants, by contrast, average 96.1\%, roughly nine times the best model's accuracy. Increasing reasoning effort barely narrows this gap.

Tool use narrows but does not close the gap. We evaluate Codex and two Claude Code agents, which write and run their own vision code and reach only 24.7--50.6\% accuracy. Gains concentrate where tasks admit reliable code-based solutions, but traversal remains difficult. On realistic imagery, models often miss tool failures. Tool use therefore shifts the active-observation bottleneck to verification in the agent loop.

In summary, we (i)~introduce \bench{}, a controlled benchmark that isolates active observation through 17 diverse tasks spanning three cognitive axes (\S\ref{sec:benchmark}); (ii)~systematically evaluate frontier MLLMs across their full reasoning-effort range and, as a tool-use test, autonomous coding agents built on them, finding a large and consistent human--model gap that neither additional reasoning nor tool use closes (\S\ref{sec:evaluation}); and (iii)~trace the residual failures to a perceptual bottleneck unresolved by additional reasoning or tooling, positioning active vision as a distinct capability that current models lack but real-world applications require (\S\ref{sec:analysis}).

\section{Related Work}

\bench{} draws on three lines of work: a long-running argument from cognitive science and computer vision that perception is intrinsically active, evidence that MLLMs struggle with basic visual perception, and existing benchmarks constructed to stress vision rather than language. We elaborate on each below.

\subsection{Active Vision in Cognitive Science and Computer Vision}
\emph{Eye Movements and Vision}~\citep{yarbus} showed that the same painting elicits different scanpaths under different task prompts, establishing that gaze is allocated by the task rather than the image alone. Later works~\citep{active_vision,active_perception} formalized the consequence in computer vision: inverse problems such as shape from shading, structure from motion, and optical flow are ill-posed for a passive observer but become well-posed once the observer can actively control its sensors as part of inference.

\emph{Animate vision}~\citep{animate_vision} reframed the same argument in terms of computational economy, using gaze-targeted local computation rather than global computation. The \emph{sensorimotor account}~\citep{sensorimotor} advanced the strongest version of this position, arguing that seeing \emph{is} the mastery of how visual input changes under one's own movements. Eye-tracking of everyday tasks supports this account beyond the laboratory~\citep{hayhoe_natural,findlay_gilchrist}. The shared prediction is sharp: without iterative sensor redirection, a vision system will fail on perceptual problems that humans solve routinely. \bench{} operationalizes that prediction for multimodal large language models.

\subsection{Vision-Centric Benchmarks for MLLMs}
A first line of work set out to isolate visual perception from linguistic priors. Early diagnoses found frontier models leaning on memorized priors and overlooking queried visual details~\citep{vlm_biased,zhang2025mllms}, and benchmarks such as MMVP~\citep{eyeswideshut}, MMStar~\citep{mmstar}, and CV-Bench~\citep{cambrian} were built to exclude items solvable from text alone and reward reading the image itself, alongside a parallel push on vision-centric model design~\citep{mm1,prismatic,eagle,brave,zhang2024euclid}. These benchmarks are now largely saturated, indicating that current MLLMs can often read the image when linguistic shortcuts are controlled, while leaving their remaining perceptual limitations unresolved.

A second line instead targets the human--model gap directly. BLINK~\citep{blink} collects tasks a person answers in the blink of an eye, BlindTest~\citep{blind} poses questions so simple that a sighted human never misses them, ArtQA~\citep{tao2024does} asks painting questions that require looking, and BabyVision~\citep{babyvision}, ChildBench~\citep{childbench}, and KidGym~\citep{kidgym} draw on tasks an infant or child can already solve. What these share is a framing by \emph{human-model gap} alone: they establish that models fall short of people without isolating \emph{which} capability is missing; we recognize active vision as a major part of that missing capability, one that deserves to be measured on its own terms. 

A third line is application-driven: examples include OSWorld~\citep{osworld2} for real-world computer-use tasks and BenchCAD~\citep{benchcad} for parametric CAD generation. These usefully track practical deployment value, but each task bundles coding, reasoning, and visual perception together, so a failure cannot be attributed to perception. \bench{} isolates exactly this capability. It targets \emph{active observation}, the iterative looking loop that prior benchmarks do not explicitly name. Its tasks resist shortcuts from any single language description and are rendered photorealistically to resemble real-world inputs encountered in downstream applications.

\section{\bench{}}
\label{sec:benchmark}

\begin{figure}[t]
  \centering
  \includegraphics[width=\linewidth]{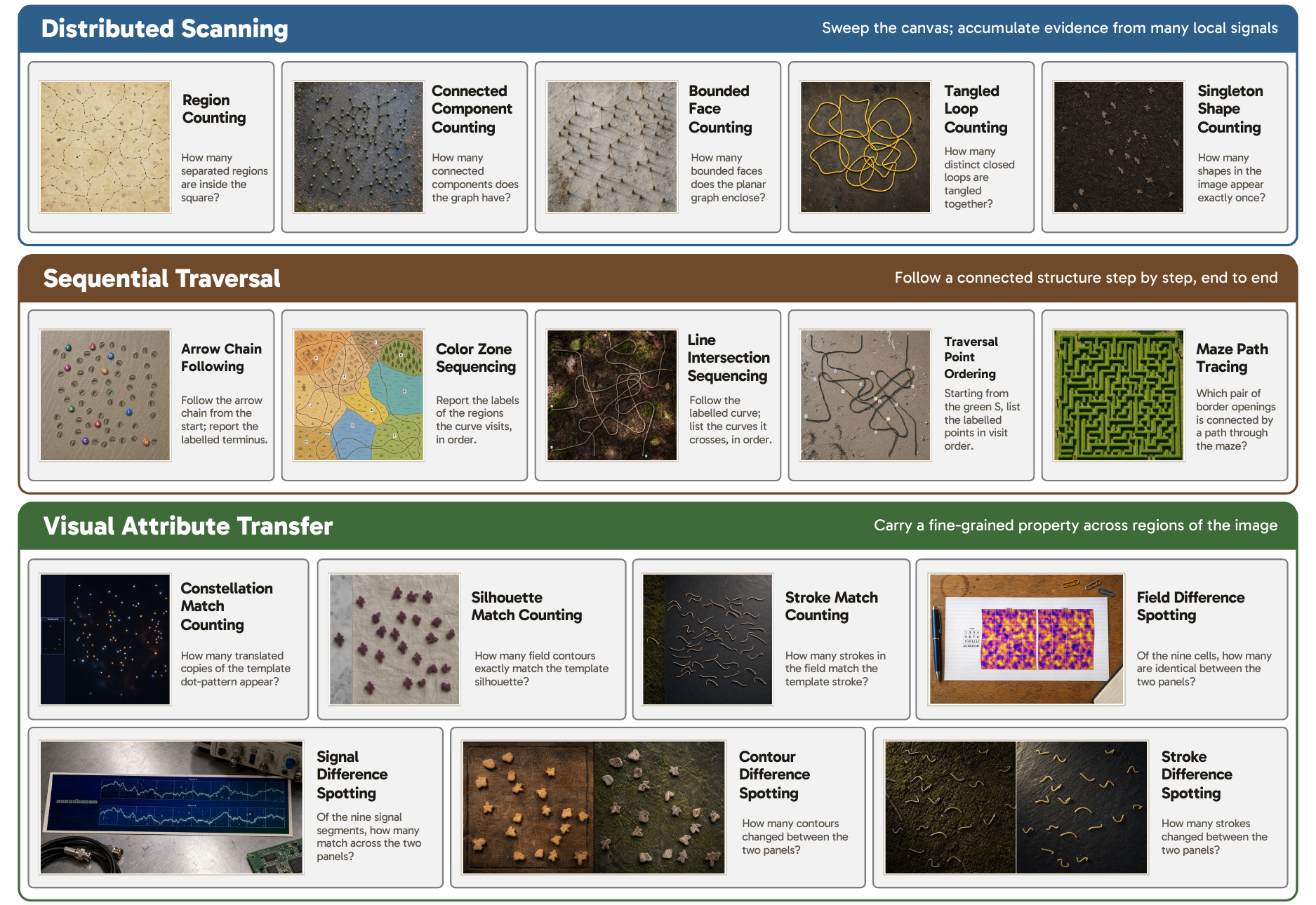}
  \caption{Overview of all 17 tasks in \bench{}, grouped into three task families that probe distinct dimensions of active observation. For each task, we show one photorealistic instance produced by the rendering pipeline described in \S\ref{sec:tasks}, together with a brief description of the required operation.}
  \label{fig:dataset-overview}
  \vspace{-0.8em}
\end{figure}

\bench{} collects tasks where a single one-shot glance is unlikely to suffice and the answer benefits from continuously returning to the pixels. \S\ref{sec:taxonomy} grounds the design in three elemental operations of human vision, following the definitions of three task families that instantiate them in Figure~\ref{fig:dataset-overview}. \S\ref{sec:antidesc} states the design principle every generator follows: discriminative visual state whose information content exceeds what one language description can losslessly carry. \S\ref{sec:tasks} introduces each of the 17 tasks and describes the photorealistic rendering pipeline that turns each procedural scaffold into the natural-looking image models actually see.

\subsection{Task Families: Grounding the Cognitive Foundations}
\label{sec:taxonomy}

\bench{} is built around three \emph{elemental operations}. Specifically, \emph{exhaustive enumeration} beyond the subitizing range~\citep{kaufman_subitizing,trick_pylyshyn}, \emph{curve tracing} along contours~\citep{jolicoeur_tracing,roelfsema_v1}, and \emph{fine-grained comparison} under visual working memory limits~\citep{luck_vogel,ballard_jit}. Psychophysics and neurophysiology characterize these operations as serial, attention-demanding routines and are carried out \emph{actively} through repeated shifts of gaze~\citep{ullman_routines,roelfsema_elemental}. We instantiate them below as three task families, each targeting a distinct dimension of active observation.

\textbf{Distributed Scanning.} \hspace{2mm}
\label{sec:ds}
The image contains many spatially distributed local signals, including dots, strokes, regions, and graph faces. Each must be found and accumulated. Difficulty scales with the number of signals and how evenly they cover the canvas. The characteristic failure has two forms. In \emph{partial coverage}, the model counts only five or six of ten items and stops before completing a full scan. In faulty \emph{individuation}, it fails to recognize each signal as a discrete element or distinguish adjacent, similar signals. Items may then be merged, split, or confused with the background before counting begins.

\textbf{Sequential Traversal.} \hspace{2mm}
\label{sec:st}
The image encodes a connected structure, such as an arrow chain, a tangled curve through colored regions, or a winding tube. The model must follow it step by step while maintaining its current position, direction, and running tally. Difficulty scales with path length, crossing density, and the visual similarity of decoys to the correct next step. The characteristic failure is \emph{gestalt interpolation}. The model guesses the endpoint from the start without traversing the intermediate steps.

\textbf{Visual Attribute Transfer.} \hspace{2mm}
\label{sec:vat}
Fine-grained comparison across regions. The model extracts a visual property from a reference region. This may be length, curvature, thickness, color arrangement, dot pattern, or orientation. It then matches or compares that property against candidates elsewhere in the image. Difficulty scales with the subtlety of the attribute distinction and the number of candidates. The characteristic failure is \emph{prior substitution}. Instead of measuring both regions, the model applies a learned linguistic prior.

\subsection{Task Design Principle: Forcing Iterative Perception}
\label{sec:antidesc}

\bench{} is built around a single design principle: every task instance carries discriminative visual state whose information content exceeds what a single language description can losslessly carry. An observer that compresses the image once (e.g., \emph{``six red circles in the top-left, three blue squares in the bottom-right''}) and then reasons over that summary will, by construction, lose the information the answer depends on; solving the task requires keeping the image itself in the loop.
The principle is realized via three properties as follows.

We realize this principle through three design properties: 
\par\vspace{-0.3\baselineskip}
\begin{itemize}[
    leftmargin=*,
    labelsep=0.5em,
    topsep=0pt,
    itemsep=1pt,
    parsep=0pt,
    partopsep=0pt
]
    \item \textbf{Arbitrary positions.} Items are placed at continuous, sampled coordinates rather than on a grid or at named anchors. Twenty dots scattered over the canvas carry 190 pairwise spatial relations and twenty real-valued coordinate pairs, far beyond what any single linguistic summary can capture. 
    
    \item \textbf{Arbitrary shapes.} Region boundaries, contours, blob silhouettes, and tile-motif outlines are synthesized fresh for every instance rather than drawn from a fixed library of named shapes. Each is either a closed contour whose radius is modulated by a random number of Fourier harmonics with sampled amplitudes and phases or a periodic spline through jittered ring waypoints. Consequently, the silhouette space is continuous and high-variance, and no two instances repeat a shape. Each boundary is defined by its exact polyline rather than by a named shape such as a triangle.
    
    \item \textbf{Arbitrary traces.} The routes the tasks ask the observer to follow, including paths, arrow chains, tangled loops, and the connecting curve through color zones, are smooth random splines through sampled control points, with dozens of meaningful inflection points that no single description preserves.
\end{itemize}

Together, these properties make iterative perception the only natural solution path. Each instance carries more visual state than a concise description can preserve, so the tasks remain readily tractable for human observers who can revisit the image yet consistently difficult for a one-pass observer to solve reliably.

\subsection{Task Instantiation: From Synthetic Scaffolds to Real-world Images}
\label{sec:tasks}

We instantiate this design with 17 task generators grouped by the three families above. Each produces a synthetic scaffold, question, and ground-truth answer from a deterministic seed, making instances reproducible. Answers span broad ranges with flat per-task distributions, preventing the modal answer or task identity from serving as a reliable shortcut. Table~\ref{tab:tasks} summarizes the resulting task set.

\begin{table}[t]
  \centering
  \footnotesize
  \setlength{\tabcolsep}{6pt}
  \setlength{\belowcaptionskip}{0pt}
  \renewcommand{\arraystretch}{1.2}
  \caption{The 17 task generators in \bench{}, grouped by cognitive axis. Each row gives the task name and a one-line description of what the model is asked to do.}
  \label{tab:tasks}
  \begin{tabular}{@{} l >{\raggedleft\arraybackslash}p{0.62\linewidth} @{}}
    \toprule
    \multicolumn{2}{@{}l@{}}{\textbf{Distributed Scanning}} \\
    \task{Bounded Face Counting}         & Count bounded faces in a planar graph drawing. \\
    \task{Connected Component Counting}  & Count connected components in a graph of dots and edges. \\
    \task{Region Counting}               & Count separated color regions inside a Voronoi-like partition. \\
    \task{Singleton Shape Counting}      & Count blob silhouettes that appear exactly once among many duplicated others. \\
    \task{Tangled Loop Counting}         & Count distinct closed loops in a tangle of curves. \\
    \addlinespace[3pt]
    \midrule
    \multicolumn{2}{@{}l@{}}{\textbf{Sequential Traversal}} \\
    \task{Arrow Chain Following}         & Follow the chain from a marked arrow; report the labeled terminus. \\
    \task{Traversal Point Ordering}      & List the labeled points encountered along the curve. \\
    \task{Color Zone Sequencing}         & List, in order, the color regions a curve visits. \\
    \task{Line Intersection Sequencing}  & List, in order, the labels of the curves crossed by the target curve. \\
    \task{Maze Path Tracing}             & Identify which pair of border openings is connected by a path. \\
    \addlinespace[3pt]
    \midrule
    \multicolumn{2}{@{}l@{}}{\textbf{Visual Attribute Transfer}} \\
    \task{Constellation Match Counting}  & Count copies of a template dot-pattern in a larger field. \\
    \task{Silhouette Match Counting}     & Count field silhouettes that are exact translated copies of a template. \\
    \task{Stroke Match Counting}         & Count field strokes that are exact translated copies of a template stroke. \\
    \task{Contour Difference Spotting}   & Count silhouette pairs whose shape differ between two paired panels. \\
    \task{Field Difference Spotting}     & List the indexed cells that differ between two paired field-grid panels. \\
    \task{Signal Difference Spotting}    & List the indexed segments that differ between two paired 1-D-signal panels. \\
    \task{Stroke Difference Spotting}    & Count stroke pairs whose shape differ between two paired panels. \\
    \bottomrule
  \end{tabular}
\end{table}

These tasks are not merely abstract puzzles. Each isolates a perceptual operation that people use in everyday life and across practical settings. For example, \task{Region Counting} resembles counting countries on a political map or zones in a thermal scan, while \task{Signal Difference Spotting} mirrors comparing ECGs or reviewing seismic traces. 
Other tasks capture topological reasoning, path tracing, and attribute matching relevant to wiring inspection, floor-plan analysis, microscopy, manufacturing inspection, and satellite change detection.

\begin{figure}[t]
  \centering
  \includegraphics[width=.87\linewidth]{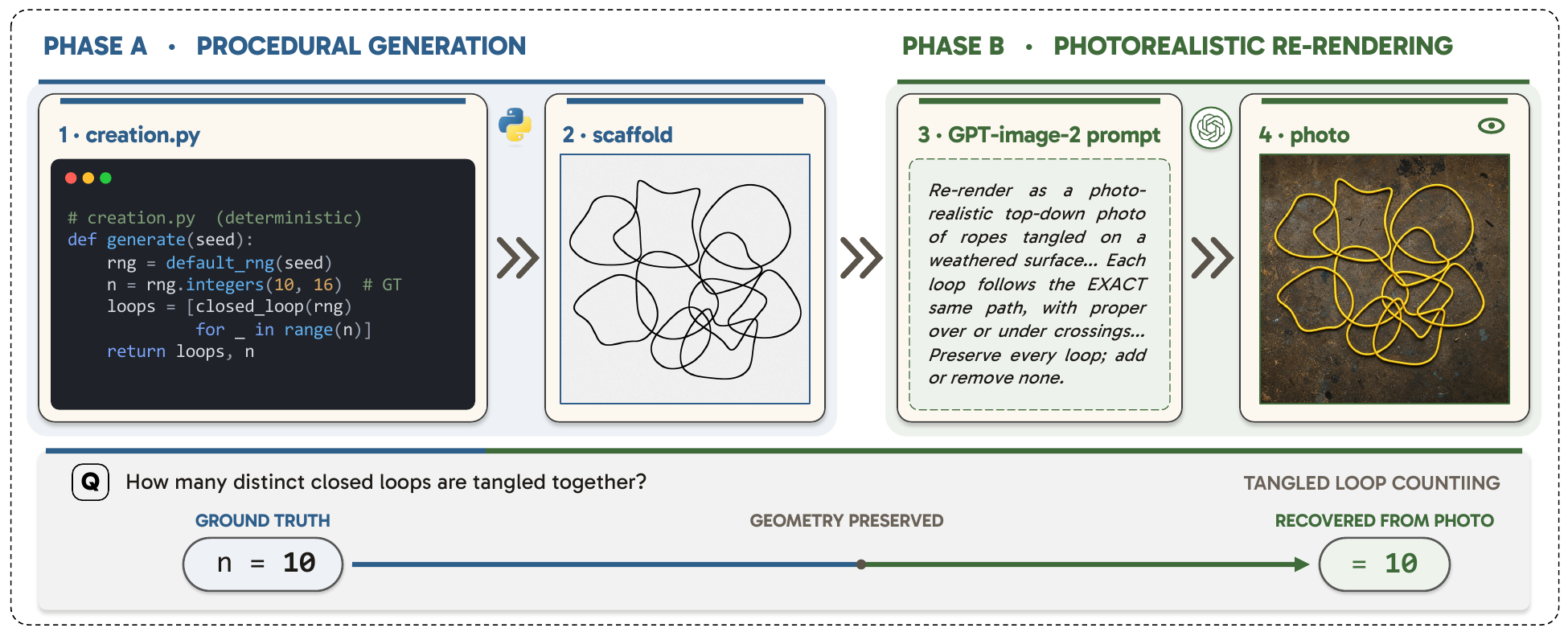}
  \caption{The four-stage generation pipeline, illustrated on \task{Tangled Loop Counting}. A deterministic Python generator (1) emits the geometric scaffold (2): random smooth closed curves with full ground truth. A task-specific GPT-image-2 prompt (3) re-renders the scaffold into a photorealistic image (4) without altering positions, counts, or topology. The benchmark question and ground-truth answer apply unchanged to both renderings; only the photorealistic image is served to the model at evaluation time.}
  \label{fig:pipeline}
\end{figure}

We place these operations in realistic settings via a two-stage pipeline (Figure~\ref{fig:pipeline}).
Each task generator first produces a procedural Matplotlib scaffold with full ground truth attached. To make the images noisy and realistic rather than cartoon-clean, we re-render every scaffold with GPT-image-2 using a task-specific prompt that maps the procedural primitives to a real-world setting. Only the rendered image is shown to the model.

The pipeline serves three purposes. First, it removes the cartoon-input confound: the perceptual difficulty seen by models is dominated by the task's discriminative structure rather than by an unfamiliar rendering style. Second, the photorealistic regime is where classical computer-vision baselines degrade sharply, in line with their known brittleness on scanned documents, ultrasound speckle, low-resolution phone snapshots, and similar real-world inputs, so the same images that test MLLMs also bound what a tool-using script can recover. Third, the imagery resembles what downstream applications actually face, making the active-vision diagnosis externally meaningful rather than an artifact of toy renderings.

\section{Benchmark Evaluation}
\label{sec:evaluation}

We first report the headline accuracy of frontier MLLMs on \bench{} (\S\ref{sec:main-result}), then dive deeper with two ablation studies on the effect of reasoning effort (\S\ref{sec:reasoning-analysis}) and of agentic tool use (\S\ref{sec:tools-analysis}).

From each of the 17 generators, we sample a fixed evaluation split of $N=5$ instances, yielding $85$ items in total. Every instance passes through the photorealistic rendering pipeline described in \S\ref{sec:tasks}. 
We use exact-match accuracy as the primary metric. Unless stated otherwise, evaluations use pure CoT with no external tools or code execution. Each item is presented as a single user message containing the question verbatim from the released manifest, followed by the image. No system prompt is used. The model is instructed to wrap its final answer in \texttt{<answer>} tags. 

We score the last such block by exact match after normalizing case, whitespace, and separators. The human baseline comprises three participants, each of whom completed the full 85-item split unaided through a self-paced web interface and was scored by the same criterion. Table~\ref{tab:main} reports frontier models at their highest reasoning effort, while Figure~\ref{fig:performance-cost} reports the full reasoning-effort range for GPT-5.5~\citep{gpt55}, Claude Fable~5~\citep{claude48}, Claude Opus~4.8~\citep{gemini35flash}, Gemini~3.1 Pro~\citep{gemini31}, and Gemini~3.5 Flash~\citep{gemini35flash}.

\begin{table}[t!]
  \centering
  \footnotesize
  \setlength{\tabcolsep}{2.4pt}
  \renewcommand{\arraystretch}{1.2}
  \caption{Per-task exact-match accuracy on \bench{}. Model entries report correct responses out of five. Human entries report the mean across three participants on the same scale (e.g., 14/15 is reported as 4.7/5). The Overall row reports scores out of 85; individual human accuracies are 97.6\%, 96.5\%, and 94.1\%.}
  \label{tab:main}
  \newcommand{\modelhead}[2]{\shortstack[c]{\vphantom{Opus 4.7}#1\\\emph{#2}}}
  \newcommand{\raisedmodelhead}[2]{\shortstack[c]{\vphantom{Opus 4.7}\raisebox{0.5ex}{#1}\\\emph{#2}}}
  \begin{tabular*}{\textwidth}{@{\extracolsep{\fill}}lccccccc@{}}
    \toprule
    \multirow{2}{*}{Task} & GPT & \multicolumn{3}{c}{Claude} & \multicolumn{2}{c}{Gemini} & Human \\
    \cmidrule(lr){2-2}\cmidrule(lr){3-5}\cmidrule(lr){6-7}\cmidrule(lr){8-8}
    & \modelhead{5.5}{xhigh} & \raisedmodelhead{Opus 4.7}{max} & \raisedmodelhead{Opus 4.8}{max} & \raisedmodelhead{Fable 5}{max} & \modelhead{3.1 Pro}{high} & \modelhead{3.5 Flash}{high} & \modelhead{avg over}{$N{=}3$} \\
    \midrule
    \multicolumn{8}{@{}l@{}}{\textbf{Distributed Scanning}} \\
    \task{Bounded Face Counting}         & 0/5 & 1/5 & 1/5 & 1/5 & 0/5 & 0/5 & 5/5 \\
    \task{Connected Component Counting}  & 0/5 & 0/5 & 0/5 & 0/5 & 1/5 & 1/5 & 4.7/5 \\
    \task{Region Counting}               & 2/5 & 0/5 & 0/5 & 0/5 & 2/5 & 1/5 & 5/5 \\
    \task{Singleton Shape Counting}      & 1/5 & 0/5 & 0/5 & 1/5 & 0/5 & 1/5 & 4/5 \\
    \task{Tangled Loop Counting}         & 0/5 & 0/5 & 0/5 & 0/5 & 0/5 & 0/5 & 5/5 \\
    \addlinespace[1pt]
    \multicolumn{8}{@{}l@{}}{\textbf{Sequential Traversal}} \\
    \task{Arrow Chain Following}         & 1/5 & 0/5 & 0/5 & 0/5 & 1/5 & 2/5 & 5/5 \\
    \task{Traversal Point Ordering}      & 0/5 & 0/5 & 0/5 & 0/5 & 0/5 & 0/5 & 5/5 \\
    \task{Color Zone Sequencing}         & 0/5 & 0/5 & 0/5 & 0/5 & 0/5 & 0/5 & 5/5 \\
    \task{Line Intersection Sequencing}  & 0/5 & 0/5 & 0/5 & 0/5 & 0/5 & 0/5 & 5/5 \\
    \task{Maze Path Tracing}             & 0/5 & 1/5 & 0/5 & 0/5 & 0/5 & 0/5 & 5/5 \\
    \addlinespace[1pt]
    \multicolumn{8}{@{}l@{}}{\textbf{Visual Attribute Transfer}} \\
    \task{Constellation Match Counting}  & 3/5 & 0/5 & 0/5 & 0/5 & 0/5 & 0/5 & 4.3/5 \\
    \task{Silhouette Match Counting}     & 1/5 & 0/5 & 0/5 & 0/5 & 0/5 & 1/5 & 5/5 \\
    \task{Stroke Match Counting}         & 1/5 & 0/5 & 0/5 & 0/5 & 1/5 & 1/5 & 4.7/5 \\
    \task{Contour Difference Spotting}   & 0/5 & 1/5 & 0/5 & 0/5 & 0/5 & 0/5 & 5/5 \\
    \task{Field Difference Spotting}     & 0/5 & 0/5 & 0/5 & 0/5 & 0/5 & 0/5 & 4.3/5 \\
    \task{Signal Difference Spotting}    & 0/5 & 0/5 & 1/5 & 0/5 & 0/5 & 0/5 & 4.7/5 \\
    \task{Stroke Difference Spotting}    & 0/5 & 1/5 & 0/5 & 1/5 & 0/5 & 0/5 & 5/5 \\
    \midrule
    \textbf{Overall}                     & \textbf{9/85} & \textbf{4/85} & \textbf{2/85} & \textbf{3/85} & \textbf{5/85} & \textbf{7/85} & \textbf{81.7/85} \\
                                         & (10.6\%) & (4.7\%) & (2.4\%) & (3.5\%) & (5.9\%) & (8.2\%) & (96.1\%) \\
    \addlinespace[1pt]
    Avg effort / item                    & 22.5k tok & 9.0k tok & 5.5k tok & 15.4k tok & 16.8k tok & 17.5k tok & 33.6\,s \\
    \bottomrule
  \end{tabular*}
\end{table}

\subsection{Main Results: Every Frontier Model Fails Most Tasks}
\label{sec:main-result}

Table~\ref{tab:main} reports per-task accuracy for six frontier models at their highest effort settings alongside a three-participant human baseline. The highest-scoring model, GPT-5.5, solves only 10.6\% (9 of 85); even Claude Fable~5, despite its strong performance on reasoning and coding leaderboards, solves just 3.5\%. On 11 of 17 tasks, GPT-5.5 scores \textbf{zero}. The human mean is 96.1\%, roughly nine times the accuracy of the best model.
The six models have only weakly overlapping success sets. No item is solved by all six, suggesting that the gap is not specific to a single model. As a shortcut check, a GPT-5.5 question-only control with the image omitted solves only 2 of 85 items, for an accuracy of 2.4\%. This result matches the run with both image and question at the same \emph{none} effort, indicating that prompt priors alone do not explain the reported performance.

\subsection{Does More Reasoning Help?}
\label{sec:reasoning-analysis}

\begin{figure}[t]
  \centering
  \includegraphics[width=0.7\linewidth]{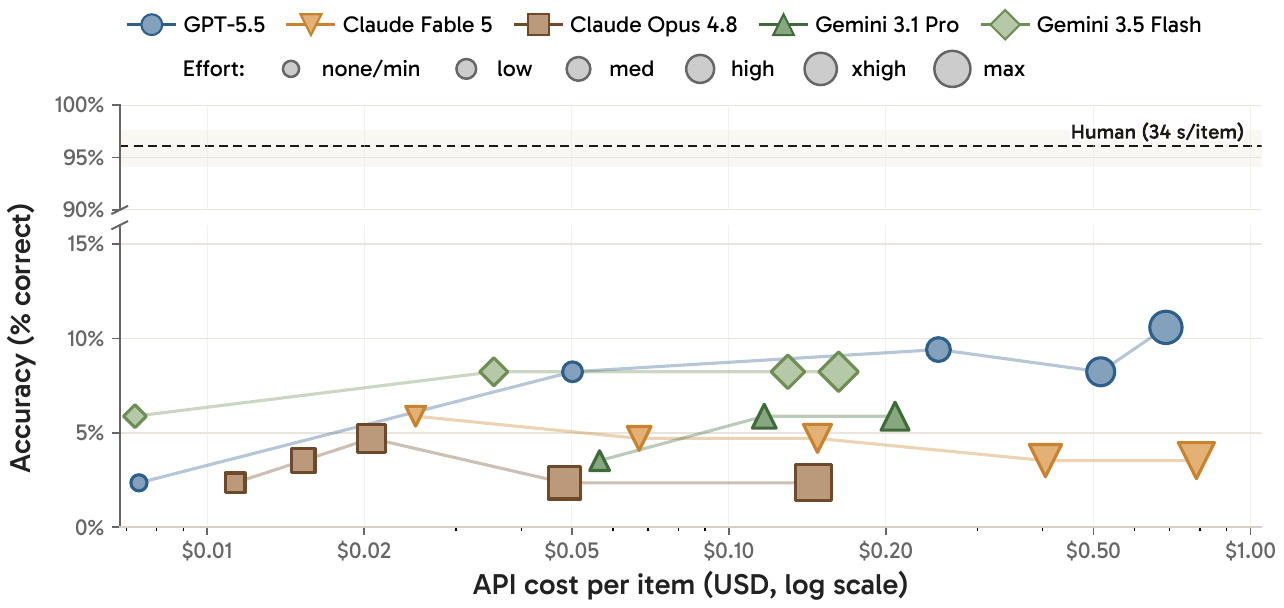}
  \caption{Accuracy versus API cost per item (log scale, public list prices) on the 85-item \bench{} split. Points are complete pure-CoT runs, with marker size encoding the reasoning-effort tier; lines connect runs of the same model. The dashed line and band are the human mean and range ($N{=}3$).}
  \label{fig:performance-cost}
\end{figure}

\begin{figure}[t]
  \centering
  \begin{minipage}[t]{0.58\linewidth}
    \vspace{0pt}
    \centering
    \includegraphics[width=\linewidth]{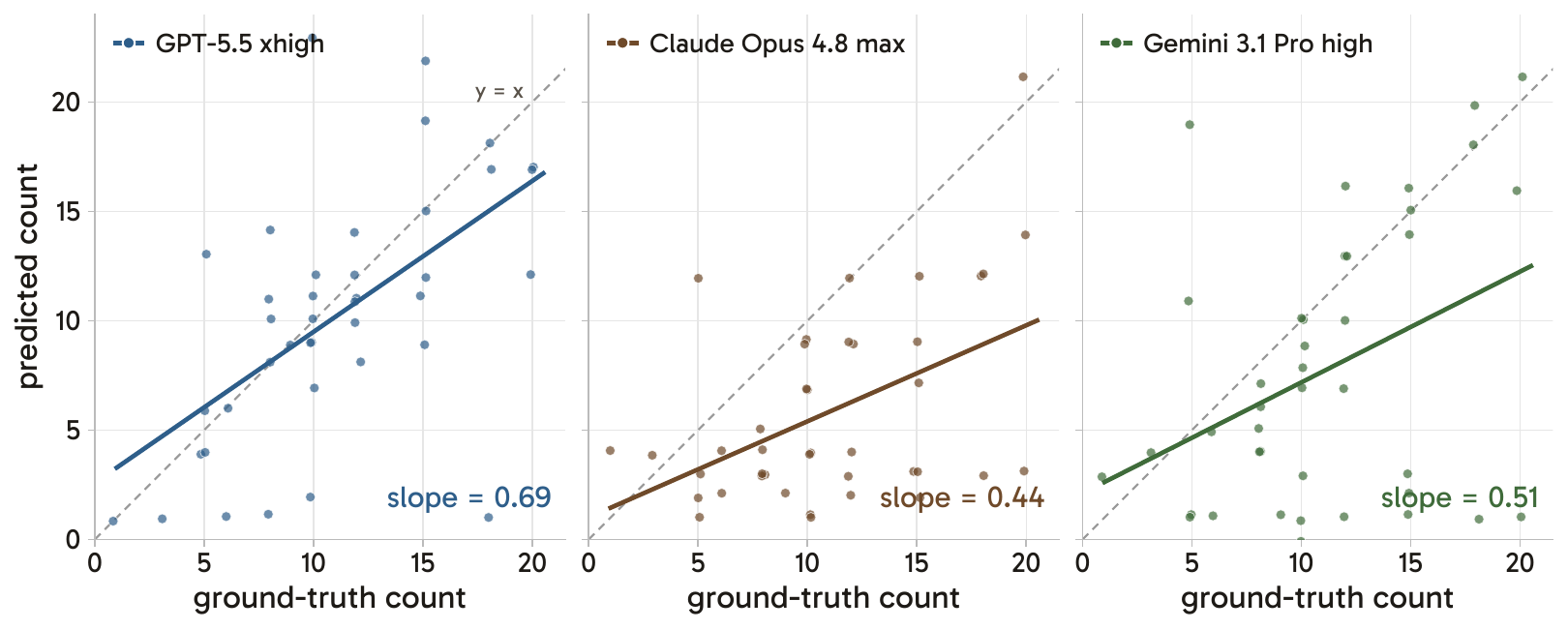}
    \caption{Predicted versus ground-truth counts across eight counting tasks at each model's highest reasoning effort. Solid lines are least-squares fits; gray dashed lines show the identity $y=x$. Slopes below one reveal increasing undercounting as the true count grows.
    }
    \label{fig:undercount}
  \end{minipage}\hfill
  \begin{minipage}[t]{0.4\linewidth}
    \vspace{0pt}
    \centering
    \includegraphics[width=\linewidth]{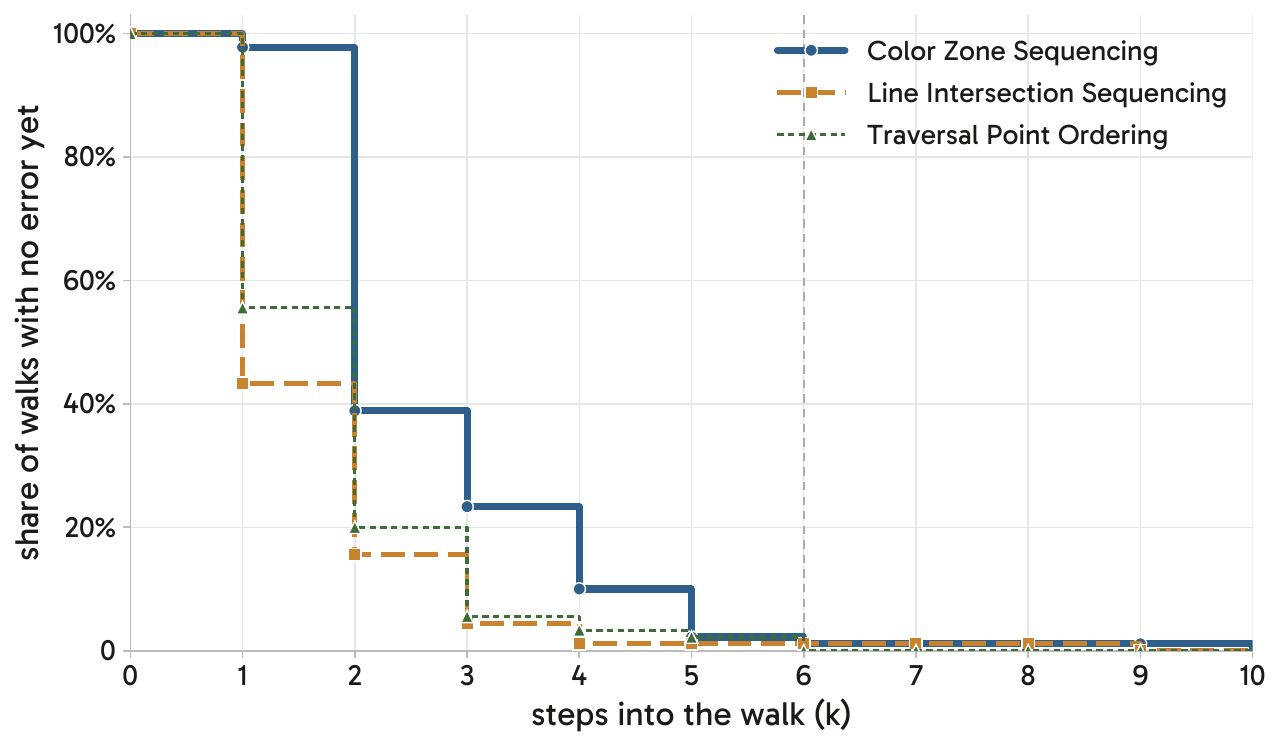}
    \caption{Prefix survival on three ordered-walk tasks (over 18 pure-CoT runs): the share of walks whose first $k$ steps are all correct. Dashed line: shortest walk length ($k{=}6$); no walk is completed exactly.}
    \label{fig:walk-survival}
  \end{minipage}
\end{figure}

The gap is not a deliberation deficit. Figure~\ref{fig:performance-cost} plots accuracy against API cost per item for every pure-CoT run across reasoning-effort tiers. Every model remains in a narrow low-accuracy band, far below the human baseline. Scaling GPT-5.5's reasoning effort from \emph{none} to \emph{xhigh} increases its per-item cost nearly a hundredfold but improves accuracy only from 2.4\% to 10.6\%. Task-level failures persist across tiers.
The other frontier models show the same pattern. Spending $31\times$ more per item on Fable~5 does not make it more accurate, and differences between effort tiers are within sampling noise on 85 items. The models do not run out of reasoning steps; they fail to pull the right visual evidence out of the image.

\subsection{What Do Models Do When They Fail?}
\label{sec:error-anatomy}

Pooling the 18 pure-CoT runs of the evaluation sweep, we examine the response behavior of each task family: what the models count, how far they trace, and when they declare a difference.

\textbf{Counting falls behind as scenes grow crowded.}\hspace{2mm}
Figure~\ref{fig:undercount} plots predicted count against ground-truth count on the eight counting tasks, with a least-squares fit per model. All fitted slopes fall well below the perfect-counting diagonal, and the fits drop further below it as the true count grows. The models are conservative counters: the more there is to count, the larger the share left uncounted, as if the model glimpses the image rather than scanning it exhaustively.

\textbf{Tracing loses its way at the very start.}\hspace{2mm}
Figure~\ref{fig:walk-survival} shows, for the three ordered-walk tasks, the share of walks whose first $k$ steps are all correct. Survival collapses within the first one or two moves, and not a single walk in the pool is completed exactly. The models cannot reliably trace along a structure in the image; failure is not gradual drift late in a long walk but the loss of the walk's frame at its very beginning.

\textbf{``Same'' becomes the safe default in visual comparison.}\hspace{2mm}
Table~\ref{tab:diff-rates} reports per-run error rates on the two panel-naming difference tasks: the share of real differences missed and the share of identical panels falsely flagged. Miss rates are high across the board while false alarms stay low, and the highlighted runs answer ``none'' on nearly every item, missing all real differences. Unable to resolve the fine-grained comparison, the models fall back on ``same'' because it is the safe answer, a response bias rather than an act of perception.

\begin{table}[t]
  \centering
  \scriptsize
  \setlength{\tabcolsep}{2pt}
  \renewcommand{\arraystretch}{1.15}
  \caption{Per-run miss and false-alarm rates on the difference tasks. Highlighted runs answer ``none'' on nearly every item, missing all real differences.}
  \label{tab:diff-rates}
  \begin{tabular*}{\textwidth}{@{\extracolsep{\fill}}l*{18}{c}@{}}
    \toprule
    \multirow{3}{*}{Rate}
      & \multicolumn{5}{c}{GPT}
      & \multicolumn{6}{c}{Claude}
      & \multicolumn{7}{c}{Gemini} \\
    \cmidrule(lr){2-6}\cmidrule(lr){7-12}\cmidrule(lr){13-19}
      & \multicolumn{5}{c}{5.5}
      & Opus 4.8 & \multicolumn{5}{c}{Fable 5}
      & \multicolumn{3}{c}{3.1 Pro} & \multicolumn{4}{c}{3.5 Flash} \\
    \cmidrule(lr){2-6}\cmidrule(lr){7-7}\cmidrule(lr){8-12}\cmidrule(lr){13-15}\cmidrule(lr){16-19}
      & \emph{none} & \emph{low} & \cellcolor{CalGoldHex}\emph{med} & \cellcolor{CalGoldHex}\emph{high} & \cellcolor{CalGoldHex}\emph{xhigh}
      & \emph{max} & \emph{low} & \emph{med} & \emph{high} & \emph{xhigh} & \emph{max}
      & \emph{low} & \emph{med} & \emph{high}
      & \emph{min} & \emph{low} & \emph{med} & \emph{high} \\
    \midrule
    Missed
      & 73\% & 96\% & \cellcolor{CalGoldHex}100\% & \cellcolor{CalGoldHex}100\% & \cellcolor{CalGoldHex}100\%
      & 81\% & 96\% & 88\% & 85\% & 77\% & 88\%
      & 96\% & 96\% & 83\%
      & 73\% & 94\% & 87\% & 90\% \\
    False al.
      & 21\% & 0\% & \cellcolor{CalGoldHex}0\% & \cellcolor{CalGoldHex}0\% & \cellcolor{CalGoldHex}0\%
      & 17\% & 12\% & 12\% & 9\% & 6\% & 8\%
      & 6\% & 3\% & 13\%
      & 13\% & 6\% & 9\% & 12\% \\
    \bottomrule
  \end{tabular*}
\end{table}

\begin{figure}[t]
  \centering
  \begin{subfigure}[t]{0.38\textwidth}\vskip 0pt
    \centering
    \includegraphics[width=\linewidth]{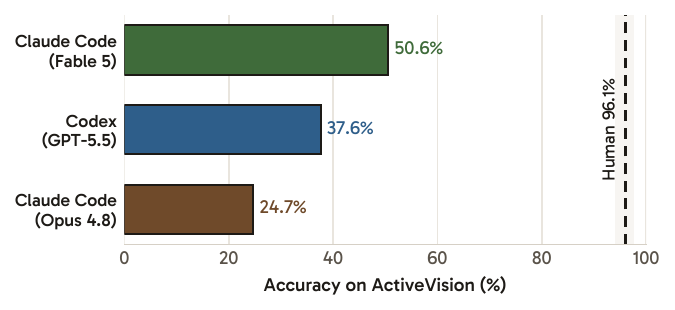}
    \caption{Per-agent accuracy: even the strongest agent (Fable~5, 50.6\%) stays far below the human band.}
    \label{fig:agent-bars}
  \end{subfigure}\hfill
  \begin{subfigure}[t]{0.6\textwidth}\vskip 0pt
    \centering
    \includegraphics[width=\linewidth]{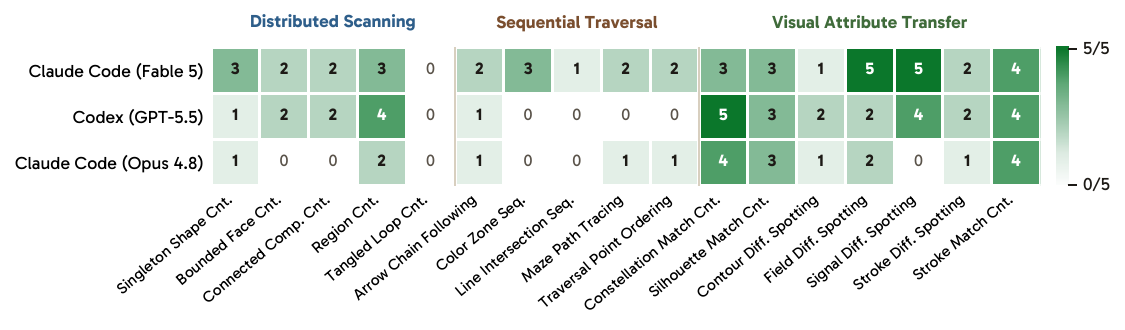}
    \caption{Per-task accuracy (correct out of 5): attribute-transfer tasks largely yield to code, the traversal family resists every agent except Fable~5, and \task{Tangled Loop Counting} resists all three agents completely.}
    \label{fig:agent-heatmap}
  \end{subfigure}
  \caption{Agentic tool use on the 85-item \bench{} split. (a)~Per-agent accuracy against the human reference ($N{=}3$); (b)~per-task breakdown across the three agents.}
  \label{fig:agents}
\end{figure}

\subsection{Can Active Vision Be Substituted by Tool Use?}
\label{sec:tools-analysis}

Our main result shows that frontier models lack active observation. We next ask whether computation can substitute for it. Code offers a strong substitution channel: brightness thresholding can segment regions, while connected-component analysis can return their count, bypassing direct visual inspection. We test this by running three autonomous coding agents per item: Codex (GPT-5.5), Claude Code (Opus~4.8), and Claude Code (Fable~5). All operate at \emph{xhigh} effort in fresh sandboxes containing only the image and question.

\begin{figure*}[!t]
    \centering
    \includegraphics[width=0.9\textwidth]{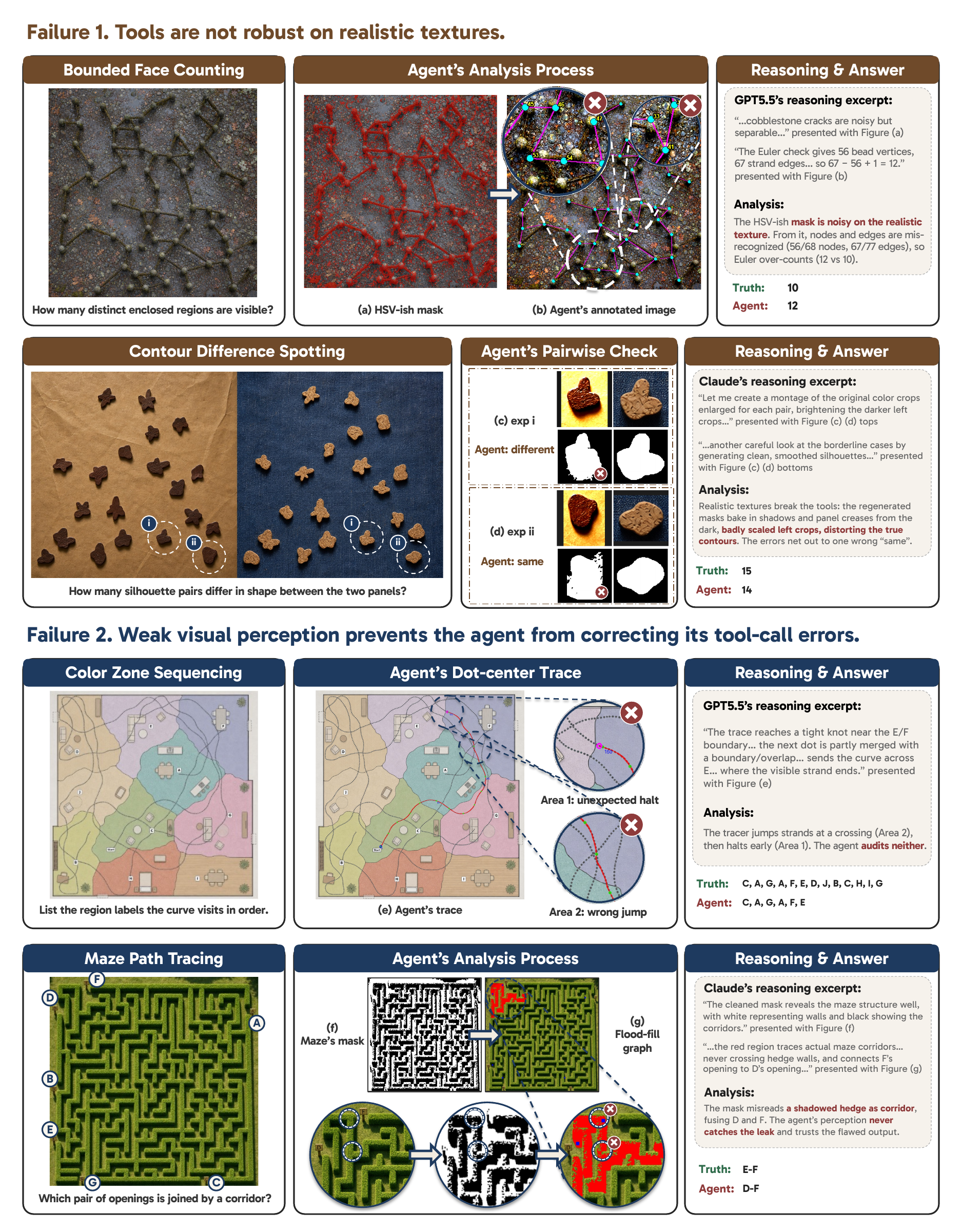}
    \vspace{-10pt} 
    \caption{Four examples illustrating two failure modes of agentic tool use, shown through the agents' own intermediate artifacts and verbatim rollout excerpts. For each failure mode, the upper example is from Codex and the lower example is from Claude Code.}
    \vspace{-10pt} 
    \label{fig:agent-failures}
\end{figure*}

\subsubsection{How Far Does Substitution Carry?}

Agentic tools improve performance substantially but do not close the human gap (Figure~\ref{fig:agent-bars}). The Fable~5 agent solves \textbf{43 of 85 items (50.6\%)}, Codex \textbf{32 (37.6\%)}, and the Opus~4.8 agent \textbf{21 (24.7\%)}, compared with 10.6\% for the best tool-free model. The gains are highly task dependent (Figure~\ref{fig:agent-heatmap}). All three agents perform best on \textbf{Visual Attribute Transfer} (43--66\%), where template cropping and measurement admit clean algorithmic reductions. In contrast, \textbf{Sequential Traversal} remains difficult: Codex solves 1/25 items, Claude Code with Opus~4.8 solves 3/25, and Claude Code with Fable~5 solves 10/25. Failure is not limited to traversal; all three agents score zero on \task{Tangled Loop Counting}. Overall, tool use helps most when perception can be reduced to a reliable computation. When no such reduction is available, the gap persists.

\begin{table}[tbp]
  \centering
  \small
  \renewcommand{\arraystretch}{1.05}
  \caption{Accuracy and mean per-item resource use on the 85-item split. Parentheses retain counts and run-level details; dashes indicate quantities that do not apply.}
  \label{tab:cost}
  \begin{tabular*}{\textwidth}{@{\extracolsep{\fill}}lrrrrrr@{}}
    \toprule
    Track
      & \multicolumn{1}{c}{\shortstack[c]{Accuracy\\(\%)}}
      & \multicolumn{1}{c}{\shortstack[c]{Time/item\\(min)}}
      & \multicolumn{1}{c}{\shortstack[c]{API cost/item\\(\$)}}
      & \multicolumn{1}{c}{\shortstack[c]{Tool calls\\per item}}
      & \multicolumn{1}{c}{\shortstack[c]{Output tokens\\per item}}
      & \multicolumn{1}{c}{\shortstack[c]{Agent turns\\per item}} \\
    \midrule
    \shortstack[l]{Codex\\{\scriptsize(GPT-5.5)}}
      & \shortstack[r]{37.6\\{\scriptsize(32/85)}}
      & \shortstack[r]{12.5\\{\scriptsize(17.8\,h total)}}
      & \shortstack[r]{2.74\\{\scriptsize(\$232.63 total)}}
      & \shortstack[r]{34\\{\scriptsize(16 code, 18 image)}}
      & \shortstack[r]{22.9k\\{\scriptsize(1.95M total)}}
      & \shortstack[r]{31\\{\scriptsize(max 84)}} \\
    \shortstack[l]{Claude Code\\{\scriptsize(Opus 4.8)}}
      & \shortstack[r]{24.7\\{\scriptsize(21/85)}}
      & \shortstack[r]{14.7\\{\scriptsize(20.8\,h total)}}
      & \shortstack[r]{4.23\\{\scriptsize(\$359.90 total)}}
      & \shortstack[r]{53\\{\scriptsize(27 code, 26 image)}}
      & \shortstack[r]{52.2k\\{\scriptsize(4.43M total)}}
      & \shortstack[r]{55\\{\scriptsize(max 115)}} \\
    \shortstack[l]{Claude Code\\{\scriptsize(Fable 5)}}
      & \shortstack[r]{50.6\\{\scriptsize(43/85)}}
      & \shortstack[r]{13.9\\{\scriptsize(19.6\,h total)}}
      & \shortstack[r]{7.63\\{\scriptsize(\$648.63 total)}}
      & \shortstack[r]{39\\{\scriptsize(20 code, 19 image)}}
      & \shortstack[r]{44.9k\\{\scriptsize(3.82M total)}}
      & \shortstack[r]{86\\{\scriptsize(max 198)}} \\
    \midrule
    \shortstack[l]{Human\\{\scriptsize($N{=}3$)}}
      & \shortstack[r]{96.1\\{\scriptsize(94.1--97.6)}}
      & \shortstack[r]{0.56\\{\scriptsize(31--69\,min total)}}
      & --- & 0 & --- & --- \\
    \bottomrule
  \end{tabular*}
\end{table}

\subsubsection{Why Does Substitution Stop There?}

Figure~\ref{fig:agent-failures} shows the two recurring failure modes. First, tools are not robust on realistic textures (Figure~\ref{fig:agent-failures}, top). On \task{Bounded Face Counting}, the agent's color mask dissolves into the cobblestone texture, so nodes and edges are mis-detected and its Euler-formula count comes out wrong (12 vs.\ a truth of 10); on \task{Contour Difference Spotting}, shadows and paper creases bake into the regenerated silhouettes and a badly scaled crop distorts the true contours, netting out to one wrong ``same.''

Second, weak perception cannot catch the tool's errors (Figure~\ref{fig:agent-failures}, bottom). On \task{Color Zone Sequencing}, the agent's tracer jumps onto a crossing strand and halts early, and the agent audits neither error, submitting the truncated route; on \task{Maze Path Tracing}, the binarized mask reads a shadowed hedge as corridor and fuses two openings, and the agent trusts the flood-fill's incorrect path. In both cases, a glance at the overlay it had just produced would falsify the answer. In short, a model that cannot see cannot tell when its code is wrong.

\subsubsection{How Costly Is Substitution?}

Substitution is also expensive and slow. Table~\ref{tab:cost} reports the per-item cost of the three tracks: each item costs an agent \$2.74--\$7.63 of compute and 12--15 minutes of wall-clock time, spent across dozens of tool calls and tens of thousands of output tokens, about 25 times longer than the half minute a human needs, plus compute a human never spends. Even the most expensive agent, Fable~5, solves only about half the benchmark; the human participants answer 96.1\% with no tools and no code.

\section{Discussion}
\label{sec:analysis}

\subsection{Interpreting the Agentic Evaluation}

\textbf{Tool-use gains and remaining bottlenecks.}\hspace{2mm}
\bench{} evaluates models as pure visual observers. The three agentic tool use runs (\S\ref{sec:tools-analysis}) show why this distinction matters. Classical computer vision primitives such as \texttt{findContours}, Canny, and template matching work best on clean inputs. On \bench{}'s photorealistic images, they fragment contours, merge objects, and fail under variations in style. A human can look back at the image and recognize that two masks belong to the same shape or that a dashed route continues beyond where the tracer stopped, while the script may return a confident but incorrect answer. Agentic tools therefore depend on active perception for robustness rather than replacing it. This challenge is greater in real deployments, where medical scans, satellite images, phone snapshots, and scanned documents contain more varied and less predictable noise than fixed pipelines can anticipate.

\textbf{Scope of the agentic evaluation.}\hspace{2mm}
The agentic ablation instead measures tool-orchestrated visual problem solving: decomposing a visual question into checkable operations, writing extraction code that works on noisy imagery, and recognizing when an output requires visual verification. In effect, the agent offloads iterative visual reasoning to code and reduces verification to simpler visual checks, a lower bar than solving the task through vision alone. Even so, the strongest agent remains far less accurate and efficient than an unaided human. Because \bench{} is organized around cognitive axes rather than real-world scenarios, fixed pipelines transfer unevenly across tasks. Figure~\ref{fig:agent-heatmap} shows this split. The attribute transfer tasks are largely solved, while the traversal tasks remain near zero for two of the three agents, and all agents fail \task{Tangled Loop Counting}. We therefore report the agentic evaluation as an ablation rather than a standalone track and leave a dedicated benchmark for agentic visual problem solving to future work.

\subsection{Limitations and Broader Implications}

\textbf{Limitations.}\hspace{2mm}
Our images are synthetic even with photorealistic re-rendering: GPT-image-2 produces realistic-looking outputs from a controlled prompt rather than samples of the natural-image distribution. We accept this in exchange for exact ground truth and controlled task structure. External validity therefore rests on the elemental visual operations the tasks measure, including scanning, tracing, and comparing, rather than on the renderings themselves (\S\ref{sec:taxonomy}). Furthermore, our tasks are built so that no short language description carries the answer; as models get better at describing images, this property can erode, and future versions of the benchmark would need to tighten the tasks to restore it.

\textbf{Outlook.}\hspace{2mm}
\bench{} reveals and quantifies a wide gap between human and machine perception that current vision-language benchmarks do not surface, turning ``active vision'' from a rhetorical claim into a number that future architectures and training objectives can be held to. The gap matters beyond the benchmark because the three operations it isolates (\S\ref{sec:taxonomy}) are the documented substance of high-stakes visual work. Exhaustive scanning underlies radiology, cell counting, inventory inspection, and aerial-image search. In chest radiography, about 30\% of missed nodules are never fixated at all~\citep{kundel_scanning}. Sequential tracing is central to connectomics, where proofreaders follow neurites through electron microscopy volumes~\citep{scheffer_hemibrain}, as well as to vessel and schematic tracing. Fine-grained comparison underlies latent-print examination, pathology, and industrial quality control~\citep{busey_fingerprint,krupinski_path,drury_inspection}. Eye-tracking across these professions shows that expertise largely \emph{is} knowing where and how to look~\citep{gegenfurtner_expertise}. A model that cannot count, trace, or compare reliably in a controlled image has no path to reliability in these settings; closing the gap is a prerequisite for the broader goal that motivates this work: reliable MLLMs and the multimodal agents built on top of them.
\section{Conclusion}
\label{sec:conclusion}

We introduce \bench{}, a benchmark that isolates active observation, the ability to keep returning to an image, forming and checking hypotheses, as reasoning unfolds. Its tasks are built so that no single language description carries the answer, and its generation pipeline preserves the exact underlying geometry while rendering it as a realistic, noisy image. In \bench{}, humans solve nearly all items in about half a minute each, whereas frontier MLLMs fail most items. Increasing reasoning effort does little to close this gap. Code and tools provide larger but uneven gains, concentrated on tasks amenable to reliable algorithmic reduction; even the strongest agent solves only half of the items while requiring far more time and compute than unaided human participants. Closing this gap is a prerequisite for many real-world applications such as robotics, healthcare, manufacturing, and scientific discovery, and \bench{} turns the gap into a number that future architectures and training objectives can be held to.%

\begingroup
\hypersetup{allcolors=black}
\bibliography{cite}
\endgroup

\end{document}